%% file: iclr2026_conference.tex
\documentclass{article} 
\usepackage{iclr2026_conference,times}

\input{math_commands.tex}

\usepackage{hyperref}
\usepackage{url}
\usepackage{booktabs}
\usepackage{float}
\usepackage{graphicx}
\usepackage{booktabs}
\usepackage{multirow}
\usepackage[table]{xcolor}
\usepackage{placeins}
\usepackage{algorithm}
\usepackage{algorithmic}  

\title{Thinking in Latents: Adaptive Anchor Refinement for Implicit Reasoning in LLMs}

\iclrfinalcopy

\author{
Disha Sheshanarayana$^{1}$ \qquad
Rajat Subhra Pal$^{2}$ \qquad
Manjira Sinha$^{2}$ \qquad
Tirthankar Dasgupta$^{2}$\\[0.45em]
$^{1}$Manipal University Jaipur\\
$^{2}$TCS Research\\[0.3em]
{\small \texttt{disha.229301161@muj.manipal.edu,}}\\
{\small \texttt{\{rajat.pal, sinha.manjira, dasgupta.tirthankar\}@tcs.com}}
}
\begin{document}

\maketitle

\begin{abstract}
Token-level Chain-of-Thought (CoT) prompting has become a standard way to elicit multi-step reasoning in large language models (LLMs), especially for mathematical word problems. However, generating long intermediate traces increases output length and inference cost, and can be inefficient when the model could arrive at the correct answer without extensive verbalization. This has motivated latent-space reasoning approaches that shift computation into hidden representations and only emit a final answer. Yet, many latent reasoning methods depend on a fixed number of latent refinement steps at inference, adding another hyperparameter that must be tuned across models and datasets to balance accuracy and efficiency. We introduce AdaAnchor, a latent reasoning framework that performs silent iterative computation by refining a set of latent anchor vectors attached to the input. AdaAnchor further incorporates an adaptive halting mechanism that monitors anchor stability across iterations and terminates refinement once the anchor dynamics converge, allocating fewer steps to easier instances while reserving additional refinement steps for harder ones under a shared maximum-step budget. Our empirical evaluation across three mathematical word-problem benchmarks shows that AdaAnchor with adaptive halting yields accuracy gains of up to 5\% over fixed-step latent refinement while reducing average latent refinement steps by 48–60\% under the same maximum-step budget. Compared to standard reasoning baselines, AdaAnchor achieves large reductions in generated tokens (92–93\%) by moving computation into silent latent refinement, offering a different accuracy–efficiency trade-off with substantially lower output-token usage.
\end{abstract}
\section{Introduction}
Large Language Models (LLMs) have demonstrated strong capabilities in mathematical reasoning, particularly when prompted to produce explicit intermediate traces such as Chain-of-Thought (CoT) \citep{wei2022cot}. Recent progress has further amplified these gains through improved instruction-tuning and longer reasoning trajectories, reinforcing the view that additional thinking tokens can unlock better problem-solving behavior \citep{kojima2022zeroshot,wang2022selfconsistency}. Despite these advances, a practical limitation persists: lengthy reasoning traces are computationally expensive, increasing decoding latency and token usage and raising serving cost especially under high-concurrency deployments. This trade-off motivates methods that preserve the benefits of multi-step reasoning while reducing the cost of token-level generation.

Several research studies have emerged to address this computational challenge. One line of research focuses on improving efficiency within the token level, for example by encouraging more concise rationales, skipping less informative tokens, or terminating early when the model appears confident in a candidate answer \citep{goyal2024pause}. While valuable, these methods remain coupled to sparse, discrete token generation and therefore inherit the cost structure of autoregressive decoding. A more promising direction explores reasoning within the dense latent space, where models can perform computation internally and emit only the final answer. Prior work has pursued latent reasoning via distillation and curriculum strategies, by reusing computation through layer skipping or looping \citep{saunshi2025looped}, or by partially replacing token-level traces with latent representations \citep{hao2024continuouslatent,zelikman2024quietstar,deng2023implicitcotkd,shen2025codi}. However, many latent reasoning methods still depend on a fixed number of latent refinement steps at inference, introducing another hyperparameter that must be tuned across models and datasets to balance accuracy and efficiency \citep{hao2024continuouslatent,deng2023implicitcotkd,deng2024explicittoimplicit,shen2025codi}.
\begin{figure}[H]
    \centering
    \includegraphics[width=\linewidth]{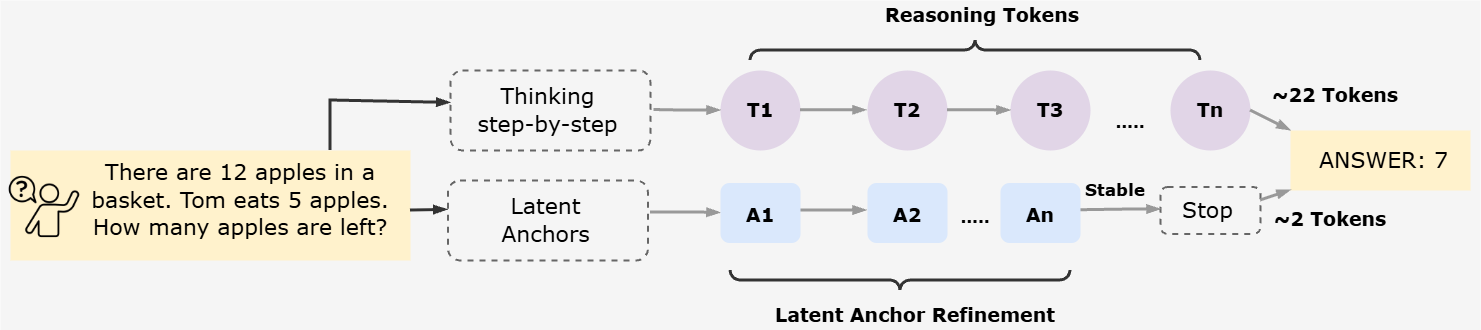}
    \caption{Comparison of AdaAnchor with explicit Chain-of-Thought (CoT) reasoning. CoT generates long intermediate reasoning tokens, whereas AdaAnchor performs implicit multi-step computation by refining latent anchor vectors and uses stability-based early stopping before answer-only decoding.}
    \label{fig:acc_vs_k}
\end{figure}
To overcome these limitations, we introduce AdaAnchor, an implicit reasoning framework that performs silent iterative computation by refining a compact set of latent anchor vectors attached to the input. Instead of generating intermediate reasoning tokens, AdaAnchor updates these anchors using the model’s hidden activations while keeping the output in an answer-only format. Furthermore, we propose an adaptive halting strategy for latent refinement: AdaAnchor tracks anchor stability across iterations and stops updating when the dynamics converge, thereby allocating computation on a per-instance basis. This enables instance-wise compute allocation under a shared maximum-step budget, so easier problems terminate quickly while harder problems receive additional refinement without requiring a fixed latent step count to be tuned per dataset \citep{graves2016act}.

We evaluate AdaAnchor on three mathematical word-problem benchmarks, GSM8K, SVAMP, and MultiArith, and compare it against fixed-step latent refinement and standard reasoning baselines under a shared maximum latent budget \citep{cobbe2021gsm8k,patel2021svamp,roy2015solving}. The results show that convergence-driven adaptive halting improves the efficiency--accuracy trade-off relative to fixed-step refinement by avoiding unnecessary latent iterations, while answer-only decoding reduces generation overhead compared to token-based baselines. Together, these findings indicate that stability-aware termination is a practical mechanism for controlling implicit computation and improving the deployability of latent reasoning methods.

\section{Related Work}

\subsection{Explicit CoT reasoning}
Large Language Models (LLMs) often exhibit substantially stronger mathematical reasoning when prompted to generate explicit intermediate traces, most notably through Chain-of-Thought (CoT) style prompting \cite{wei2022cot}. Follow-up work showed that even without exemplars, carefully designed prompts can elicit multi-step reasoning behaviors \cite{kojima2022zeroshot}, and decoding-time strategies such as self-consistency can further improve reliability by aggregating over diverse reasoning paths \cite{wang2022selfconsistency}. At the same time, generating long token-level rationales can be costly in practice, motivating methods that reduce the number of produced tokens while preserving reasoning quality. Representative directions include introducing structured pause mechanisms that encourage internal computation without fully verbalizing every step \cite{goyal2024pause}, as well as prompting frameworks that constrain or compress intermediate traces to be shorter and cheaper to decode \cite{aytes-etal-2025-sketch,xu2025softcot}. Orthogonally, improvements at the training objective level—such as predicting multiple future tokens per position have been explored as a way to increase throughput and reduce inference time without changing the basic autoregressive interface \cite{gloeckle2024better}.

\subsection{Latent reasoning}
To avoid the overhead of long textual traces, latent reasoning approaches shift more computation into hidden representations and emit only the final answer. A prominent line of work performs silent reasoning by feeding internal hidden states back into the model as continuous thoughts, enabling multi-step computation without committing to discrete tokens at every step \cite{hao2024continuouslatent}. Related approaches learn latent reasoning trajectories via distillation or self-training so that intermediate computation is represented implicitly rather than as explicit text \cite{shen2025codi}. More recently, hybrid schemes have explored mixing latent and text tokens during training and inference, aiming to retain some controllability and interpretability while still gaining efficiency from latent computation \cite{su2025tokenassorted}. In parallel, latent-space compression methods seek to reduce the length of reasoning traces by compressing multiple reasoning tokens into fewer latent steps, improving efficiency while keeping performance competitive \cite{tan2025thinksilently}. Another complementary direction analyzes or exploits latent dynamics during generation to guide or improve reasoning behavior \cite{zelikman2024quietstar}.

\subsection{Adaptive halting and compute allocation}
A recurring limitation in iterative (token or latent) reasoning is that many methods require fixing the number of refinement steps in advance, which can lead to over-computation on easy instances and under-computation on hard ones. Adaptive computation mechanisms address this by allowing models to halt dynamically based on instance difficulty, with classic formulations such as Adaptive Computation Time (ACT) learning when to stop iterating \cite{graves2016act}. 

Prior work on latent and implicit reasoning often (i) runs a fixed number of silent refinement steps, (ii) relies on task-specific training signals to control computation, or (iii) lacks explicit budget control. In contrast, AdaAnchor introduces a compact latent bottleneck via learnable anchor vectors and a convergence-based halting rule, enabling per-example adaptive compute under a shared maximum-step budget without training a separate halting controller.

\section{Method}
\subsection{Problem Formulation}
We consider mathematical word-problem solving in an answer-only setting. Given an input question $x$ (a sequence of tokens), the goal is to predict the correct final answer $y$. Standard reasoning prompting often generates an intermediate rationale $r$ before producing the answer. While such explicit traces can improve accuracy, they incur substantial inference cost because both the rationale and the answer must be generated autoregressively, increasing latency and token usage.

Our objective is to retain the benefits of multi-step reasoning while reducing the overhead of token-level generation. We work with a base autoregressive transformer $f_{\theta}$ and introduce a compact latent state in the form of anchor vectors. Let $A^{(0)} \in \mathbb{R}^{m \times d}$ denote $m$ anchor embeddings of dimension $d$. We augment the model input in embedding space by prepending projected anchor embeddings to the token embeddings, denoted $[P(A^{(t)}); Emb(x)]$. Instead of decoding a long rationale, the model performs silent iterative computation by repeatedly updating the anchor state for a variable number of refinement steps $T$, producing a sequence $\{A^{(t)}\}_{t=0}^{T}$. After refinement terminates, the model generates the final answer $\hat{y}$ conditioned on the refined anchors and the original question, while keeping the output in an answer-only format.

We measure efficiency along two axes that directly affect inference cost: (i) the number of refinement iterations used per instance, and (ii) the number of generated output tokens. The core problem is therefore to design (a) an anchor-based refinement mechanism that supports implicit multi-step computation and (b) an instance-wise stopping rule that selects $T \le K_{\max}$ automatically, so that the model avoids unnecessary refinement on easy inputs while allocating additional latent computation to harder ones. The AdaAnchor refinement procedure and the stability-based halting rule are described in Sections~\ref{sec:adaanchor} and~\ref{sec:halting}, respectively.
\begin{figure}[H]
    \centering
    \includegraphics[width=\linewidth]{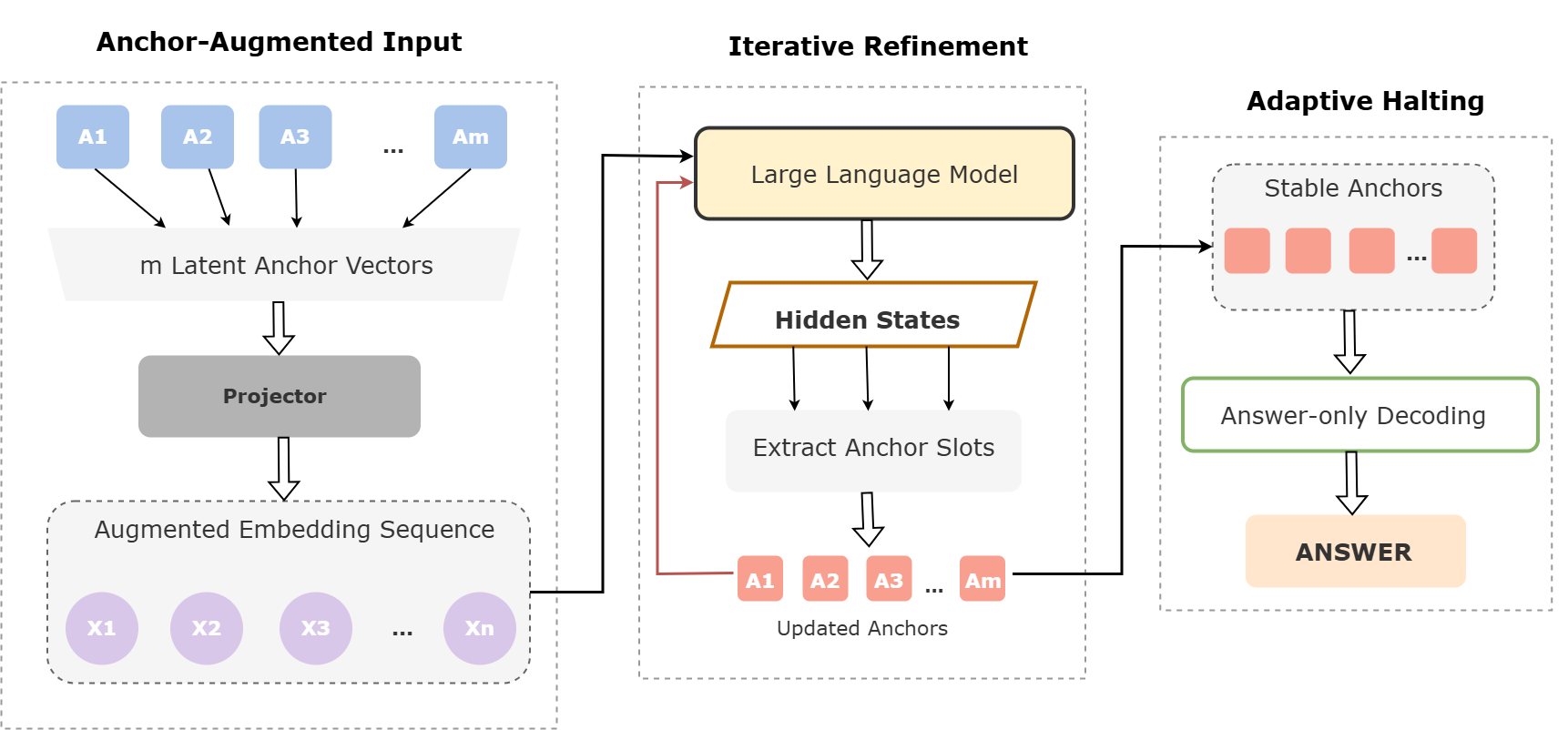}
    \caption{\textbf{Overview of AdaAnchor.} AdaAnchor prepends $m$ learnable latent anchor vectors to the input embedding sequence (left), iteratively refines them via repeated forward passes and anchor-slot updates (middle), and uses a stability-based criterion to halt early before performing answer-only decoding (right).}
    \label{fig:main}
\end{figure}
\subsection{AdaAnchor Framework}
\label{sec:adaanchor}
We propose AdaAnchor, a latent reasoning framework that enables implicit multi-step computation in large language models without generating explicit intermediate reasoning tokens (Figure~\ref{fig:main}). AdaAnchor augments a base autoregressive transformer $f_{\theta}$ with a compact set of learnable anchor vectors prepended to the input in embedding space. These anchors act as a reusable low-dimensional state that is iteratively refined through repeated forward passes, allowing the model to think silently while keeping the textual interface in an answer-only format.

\paragraph{Anchor-augmented input.}
Let the tokenized question be $x=(x_1,\dots,x_n)$, and let $\mathrm{Emb}(x)\in\mathbb{R}^{n\times d}$ denote its input embedding sequence (hidden size $d$). AdaAnchor maintains $m$ anchor vectors
\begin{equation}
A^{(t)} = [a^{(t)}_1,\dots,a^{(t)}_m]\in\mathbb{R}^{m\times d},
\end{equation}
where $t$ indexes the refinement iteration. At inference, we initialize anchors with the learned parameters, i.e., $A^{(0)} \leftarrow A_{\text{learned}}$. At iteration $t$, we form the augmented embedding sequence
\begin{equation}
E^{(t)} = [\tilde{A}^{(t)};\ \mathrm{Emb}(x)], 
\qquad \tilde{A}^{(t)} = P(A^{(t)}), 
\tag{2}
\end{equation}
where $P(\cdot)$ projects anchor vectors into the same embedding space as token embeddings.

Prior soft-prompting or prefix-tuning methods learn a set of static prompt embeddings that are prepended to the input and remain fixed throughout inference \citep{lester2021power,li2021prefix}. AdaAnchor's anchors are different: they form an explicit latent state $A^{(t)}$ that is iteratively rewritten during inference. After each forward pass on $[P(A^{(t)});\mathrm{Emb}(x)]$, we update the anchor slots using the model’s hidden states and feed the refined anchors into the next refinement step. In this way, anchors act as a persistent latent memory across refinement iterations rather than a fixed learned prefix, enabling silent multi-step computation under a shared refinement budget.

\paragraph{Iterative anchor refinement.}
Given $E^{(t)}$, we run a forward pass through the base LM to obtain
\begin{equation}
H^{(t)} = f_{\theta}(E^{(t)}) \in \mathbb{R}^{(m+n)\times d},
\tag{3}
\end{equation}
where $H^{(t)}$ denotes the final-layer hidden states for all positions in the augmented sequence. AdaAnchor updates the anchors by extracting the hidden states corresponding to the anchor positions.

When $\beta = 1$, the update reduces to overwriting; smaller $\beta$ smooths anchor evolution and improves convergence behavior in practice. The refinement loop runs for at most $K_{\max}$ iterations, and it can terminate early via the adaptive halting criterion described in Section~\ref{sec:halting}.
\paragraph{Answer-only decoding.}
After refinement terminates at step $T$, AdaAnchor generates the final answer by decoding only a short answer continuation conditioned on the refined anchors and the original input. By performing computation through latent refinement rather than token-level rationales, AdaAnchor substantially reduces generated tokens while keeping the output concise.

\begin{algorithm}[t]
\caption{AdaAnchor Inference with Adaptive Halting}
\label{alg:adaan_infer}
\begin{algorithmic}[1]
\REQUIRE Tokenized question $x$, learned anchor parameters $A_{\text{learned}}\in\mathbb{R}^{m\times d}$, max refinement budget $K_{\max}$, smoothing $\beta\in(0,1]$, threshold $\tau$, patience $s$
\STATE Set anchors $A^{(0)} \leftarrow A_{\text{learned}}$
\STATE $c \leftarrow 0$ \COMMENT{consecutive stable counter}
\FOR{$t=0,1,\dots,K_{\max}-1$}
    \STATE Form augmented input $E^{(t)} \leftarrow [\,P(A^{(t)});\ \mathrm{Emb}(x)\,]$
    \STATE Run backbone LM on $E^{(t)}$ to obtain hidden states $H^{(t)}$
    \STATE Extract anchor-position states $A_{\text{new}}^{(t+1)} \leftarrow H^{(t)}_{1:m}$
    \STATE Smooth update $A^{(t+1)} \leftarrow (1-\beta)A^{(t)} + \beta A_{\text{new}}^{(t+1)}$
    \STATE Compute stability $\Delta^{(t+1)} \leftarrow 1-\cos(\bar{a}^{(t+1)},\bar{a}^{(t)})$,
           where $\bar{a}^{(t)}=\frac{1}{m}\sum_{i=1}^{m} a_i^{(t)}$ and $a_i^{(t)}$ denotes the $i$-th anchor vector in $A^{(t)}$
    \IF{$\Delta^{(t+1)} < \tau$}
        \STATE $c \leftarrow c + 1$
    \ELSE
        \STATE $c \leftarrow 0$
    \ENDIF
    \IF{$c \ge s$}
        \STATE \textbf{break}
    \ENDIF
\ENDFOR
\STATE Decode answer-only $\hat{y}$ conditioned on $[\,P(A^{(t+1)});\ \mathrm{Emb}(x)\,]$
\RETURN $\hat{y}$
\end{algorithmic}
\end{algorithm}

\subsection{Adaptive Halting}
\label{sec:halting}
We incorporate an adaptive halting mechanism that determines how many latent refinement steps to run per instance. Since AdaAnchor performs iterative computation by updating the anchor state $A^{(t)}$, we use anchor stability across iterations as a convergence signal. Intuitively, when refinement is still productive, consecutive anchor states change noticeably; once refinement has converged, anchor updates become small and repetitive.

\paragraph{Anchor stability metric.}
We quantify refinement progress by measuring the change between successive anchor states. In particular, we summarize the anchors at iteration $t$ by their mean representation
\begin{equation}
\bar{a}^{(t)}=\frac{1}{m}\sum_{i=1}^{m} a_i^{(t)},
\tag{4}
\end{equation}
and define a scalar update magnitude via cosine distance dissimilarity:
\begin{equation}
\Delta^{(t)} = 1 - \cos\!\big(\bar{a}^{(t)},\bar{a}^{(t-1)}\big).
\tag{5}
\end{equation}
Smaller $\Delta^{(t)}$ indicates that the anchor dynamics are approaching a fixed point, suggesting that further refinement is unlikely to add useful computation. (An equivalent alternative is to average the change across all anchors; we adopt this compact summary for robustness and efficiency.)

\paragraph{Halting rule.}
AdaAnchor refines anchors for at most $K_{\max}$ iterations, but it can stop early when the anchor updates converge. Concretely, we halt at the first iteration $T$ such that the update magnitude remains below a threshold $\tau$ for $s$ consecutive steps:
\begin{equation}
T = \min \left\{ t \in \{1,\dots,K_{\max}\} \;:\; \Delta^{(t-j)} < \tau \;\; \forall j \in \{0,\dots,s-1\} \right\}.
\tag{6}
\end{equation}
This yields instance-wise compute allocation under a shared maximum-step budget: easier instances typically converge in fewer steps, while harder ones continue refining until convergence or the budget limit. The halting check introduces negligible overhead because it operates directly on anchor states already computed during refinement.

\section{Experiments}
In this section, we evaluate our method AdaAnchor on mathematical word-problem solving benchmarks and compare it against fixed-step latent refinement and standard reasoning baselines under a shared maximum latent budget.

\subsection{Experimental Setup}

\subsubsection{Datasets}
Our training is conducted primarily on \textbf{GSM8K} and \textbf{SVAMP}. Specifically, we use approximately 7.47K training examples from GSM8K and 700 training examples from SVAMP to fine-tune the models and learn the AdaAnchor components. We evaluate the trained models on three mathematical word-problem benchmarks: \textbf{GSM8K}, \textbf{SVAMP}, and \textbf{MultiArith}. The corresponding test sets contain 1.32K, 300, and 600 examples, respectively. For all datasets, we use an answer-only setup, where the model is prompted with the question and is expected to output only the final numeric answer.
\begin{table}[H]
\centering
\caption{Overview of datasets used.}
\vspace{2mm} 
\label{tab:dataset_overview}
\renewcommand{\arraystretch}{1.15}
\setlength{\tabcolsep}{10pt}
\begin{tabular}{l c c}
\hline
\textbf{Dataset} & \textbf{Task Type} & \textbf{Metric} \\
\hline
GSM8K\citep{cobbe2021gsm8k}   & Grade-school mathematical word problems & Accuracy \\
SVAMP\citep{patel2021svamp}      & Arithmetic word problems & Accuracy \\
MultiArith\citep{roy2015solving} & Multi-step arithmetic word problems & Accuracy \\
\hline
\end{tabular}
\end{table}

\subsubsection{Models}
We train and evaluate on two backbone language models, \textbf{Qwen2.5-1.5B}\citep{qwen25techreport} and \textbf{Llama-3.2-1B}\citep{grattafiori2024llama3} parameters, across all experiments. We focus on small LMs, where efficiency differences from latent refinement are easier to observe and measure reliably. Throughout our experiments, we use deterministic decoding and hold inference settings constant across methods to ensure a fair comparison.

\subsubsection{Baselines and Metrics}
\paragraph{Baselines.}
We compare AdaAnchor against token-based and latent reasoning baselines that reflect common evaluation settings for efficient reasoning.
(1) \textbf{No CoT}: the model is prompted to directly output the final answer in an answer-only format, without generating intermediate reasoning.
(2) \textbf{CoT}: the model generates an explicit step-by-step rationale followed by the final answer using standard Chain-of-Thought prompting.
(3) \textbf{iCoT}: an implicit-CoT style baseline that removes explicit reasoning traces while retaining an answer-only output format, serving as a lightweight implicit reasoning comparison.

\paragraph{Metrics.}
We evaluate using three metrics: \textbf{Accuracy} (exact-match correctness of the final answer under deterministic decoding), \textbf{Average Tokens} (average number of generated output tokens per example), and \textbf{Average Steps} (average number of latent refinement iterations executed per example).

\subsubsection{Implementation Details}

During training, we keep the backbone LM weights frozen and optimize only the AdaAnchor-specific components (learnable anchor embeddings and the small projection used to inject anchors), together with low-rank adaptation modules (LoRA) \citep{hu2021lora}. We train for 20 epochs using mixed precision. Optimization uses AdamW \citep{loshchilov2017adamw} with a fixed learning rate of $1\mathrm{e}{-4}$ and weight decay $1\mathrm{e}{-2}$, per-device batch size 1, and gradient accumulation of 16; we select the final checkpoint based on validation accuracy. In addition to the answer-only loss, we include an auxiliary anchor-alignment objective derived from coarse chunks of the rationale text, with dataset-dependent weighting as implemented in our code. At inference, AdaAnchor refines anchors up to a maximum budget using a smoothed update, and applies stability-based early stopping by monitoring anchor change across iterations and halting once it consistently converges.

\subsection{Results}
Table ~\ref{tab:main_results} compares AdaAnchor against standard baselines on three mathematical word-problem benchmarks, GSM8K, SVAMP, and MultiArith, under an answer-only evaluation setup. Across datasets, AdaAnchor consistently improves over No-CoT prompting, delivering relative accuracy gains of roughly $\sim$23--32\% on Qwen2.5-1.5B and $\sim$39--64\% on Llama-3.2-1B, while keeping output-token usage very low. In particular, compared to explicit CoT, AdaAnchor reduces generated tokens by about $\sim$90--93\%, highlighting a different accuracy--efficiency trade-off in which more computation is shifted into silent latent refinement rather than verbose token-level rationales.

Moreover, the adaptive halting variant further improves this trade-off by avoiding unnecessary refinement once anchor dynamics stabilize. Relative to a fixed refinement budget ($K=8$), adaptive stopping uses $\sim$48--61\% fewer latent refinement steps on average while maintaining similar accuracy and improving it in several cases, indicating effective instance-wise compute allocation: easier instances terminate early, while harder ones continue refining under the same maximum-step budget. Overall, these results suggest that convergence-aware latent refinement can support iterative reasoning with substantially lower output-token usage and fewer refinement steps.
\begin{table*}[t]
\centering
\small
\setlength{\tabcolsep}{5pt}
\renewcommand{\arraystretch}{1.15}

\resizebox{\textwidth}{!}{%
\begin{tabular}{l|l|ccc|ccc|ccc}
\toprule
\multicolumn{2}{c|}{  } &
\multicolumn{3}{c|}{\textbf{GSM8K}} &
\multicolumn{3}{c|}{\textbf{SVAMP}} &
\multicolumn{3}{c}{\textbf{MultiArith}} \\
\midrule
\textbf{Model} & \textbf{Method} &
\textbf{Acc.} & \textbf{Avg Tok} & \textbf{Avg Steps} &
\textbf{Acc.} & \textbf{Avg Tok} & \textbf{Avg Steps} &
\textbf{Acc.} & \textbf{Avg Tok} & \textbf{Avg Steps} \\
\midrule

\multirow{5}{*}{Qwen2.5-1.5B}
& No CoT               & 13.0  & 2.16 & --   & 42.0 & 2.34 & --   & 22.3  & 2.41 & -- \\
& CoT                  & 20.0  & 28.27& --   & 59.3 & 29.09& --   & 34.3  & 30.2 & -- \\
& iCoT                 & 12.23 & 2.36 & --   & 48.5 & 2.04 & --   & 28.56 & 1.66 & -- \\
\rowcolor{gray!15}
& \textbf{AdaAnchor ($K{=}8$)}  & 16.0  & 2.73 & 8    & 50.5 & 2.12 & 8    & 27.6  & 2.34 & 8  \\
\rowcolor{gray!15}
& \textbf{AdaAnchor (adaptive)} & 16.0  & 2.17 & 3.23 & 55.2 & 2.23 & 4.12 & 29.4  & 2.16 & 3.82 \\
\midrule
\multirow{5}{*}{Llama-3.2-1B}
& No CoT               & 10.5  & 2.98 & --   & 38.2 & 2.10 & --   & 20.56 & 2.08 & -- \\
& CoT                  & 23.2  & 25.4 & --   & 57.8 & 28.21& --   & 43.33 & 28.0 & -- \\
& iCoT                 & 11.7  & 2.25 & --   & 54.2   & 2.43 & --   & 30.84 & 2.12 & -- \\
\rowcolor{gray!15}
& \textbf{AdaAnchor ($K{=}8$)}  & 14.0  & 2.89 & 8    & 52.0 & 2.13 & 8    & 28.31 & 2.48 & 8  \\
\rowcolor{gray!15}
& \textbf{AdaAnchor (adaptive)} & 17.2  & 2.45 & 3.5  & 53.4 & 2.8  & 3.1  & 32.44 & 2.57 & 3.5 \\

\bottomrule
\end{tabular}%
}

\vspace{2mm}
\caption{Experimental results on mathematical reasoning benchmarks. We report 
accuracy (Acc. \%), average output tokens (Avg Tok), and average latent 
refinement steps (Avg Steps) for answer-only evaluation. No CoT denotes direct answer generation without reasoning. 
CoT generates full chain-of-thought rationales before the answer. iCoT 
uses implicit chain-of-thought.}
\label{tab:main_results}
\end{table*}

\subsection{Ablation Study}
We conduct an ablation study to isolate the contribution of stability-based adaptive halting in AdaAnchor. Specifically, we compare a fixed-step refinement variant that always runs a constant number of latent refinement iterations ($K=8$) against the adaptive variant that refines anchors up to the same maximum budget ($K_{\max}=8$) but halts early once anchor updates converge. Both variants use the same anchor design, refinement update rule, and decoding configuration; the only difference is whether refinement proceeds for a fixed number of iterations or terminates based on stability.

\begin{figure}[H]
    \centering
    \includegraphics[width=\linewidth]{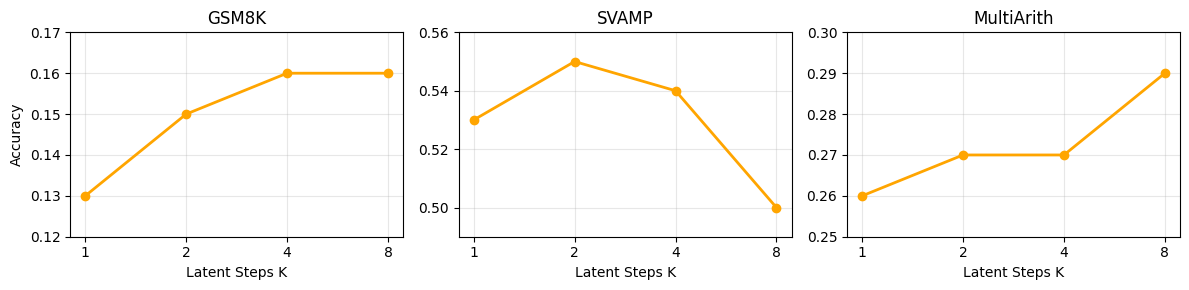}
    \caption{Accuracy vs. fixed latent refinement budget $K$ on Qwen2.5-1.5B. Each panel reports performance as a function of $K \in \{1,2,4,8\}$.}
    \label{fig:as1}
\end{figure}

To understand how performance depends on the available refinement budget, we vary the fixed-step refinement length on the Qwen2.5-1.5B using $K \in \{1,2,4,8\}$ as shown in Figure~\ref{fig:as1}. Accuracy generally improves as the budget increases, but the gains saturate beyond moderate values, indicating diminishing returns from always executing a large fixed $K$. This supports the motivation for convergence-aware termination, which can avoid unnecessary refinement when additional iterations provide limited benefit.

\begin{figure}[H]
    \centering
    \includegraphics[width=\linewidth]{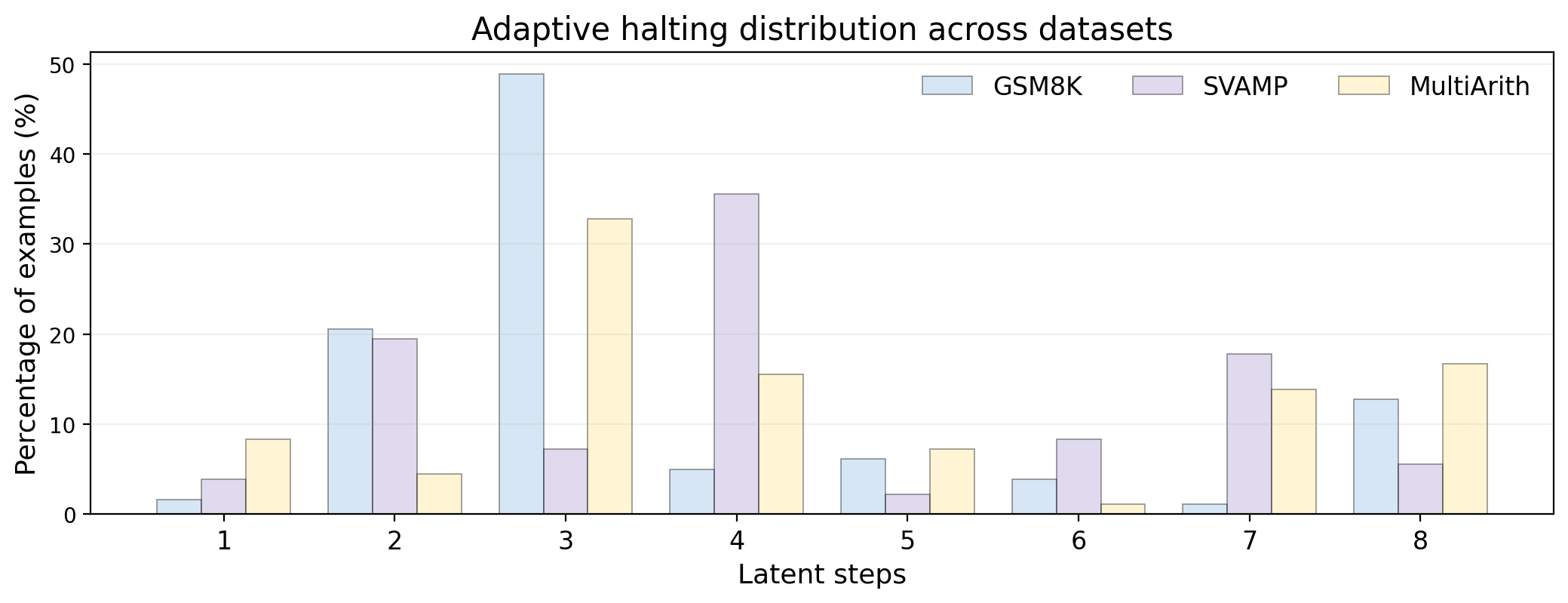}
    \caption{Adaptive halting distribution showing percentage of examples halting at each step (1-8) across datasets on Qwen2.5-1.5B.}
    \label{fig:as2}
\end{figure}

We also analyze the refinement lengths selected by adaptive halting on Qwen2.5-1.5B under the shared maximum budget $K_{\max}=8$ as shown in Figure~\ref{fig:as2}. The halting-step distribution shows that the model frequently stops well before reaching the maximum budget, while reserving more refinement steps for a smaller fraction of harder instances. This behavior confirms that the stability criterion induces instance-wise compute allocation in practice and explains the reduction in average latent steps relative to fixed-step refinement without tuning a fixed latent-step hyperparameter per dataset.

\section{Limitations and Future Work}
While AdaAnchor improves the efficiency--accuracy trade-off via stability-based adaptive halting, it has few limitations. First, our halting mechanism relies on a hand-designed stability criterion. Although this heuristic performs well in our setting, it may be sensitive to hyperparameter choices and can occasionally halt too early or too late on atypical inputs or under distribution shifts. Second, while anchor refinement provides a compact latent reasoning state, the semantics of the learned anchors are not directly interpretable, and it remains difficult to precisely attribute improvements to specific latent behaviors compared to explicit token-level rationales.

These limitations motivate several directions for future work. A natural extension is to replace the heuristic stopping rule with a learned halting policy, for example via a lightweight controller trained with supervision or reinforcement learning, or by incorporating calibrated confidence/verification signals to make termination more robust across models and datasets. In addition, improving interpretability of anchor dynamics is an important direction: future work could introduce probing and visualization tools for anchor trajectories, enforce structured anchors aligned to intermediate quantities , or add auxiliary objectives that encourage anchors to correspond to human-interpretable sub-computations.

\section{Conclusion}
We introduced AdaAnchor, a latent reasoning framework that performs multi-step computation by iteratively refining a compact set of learnable anchor vectors while keeping the model output in an answer-only format. By shifting reasoning from token-level traces into a small latent state, AdaAnchor targets lower output-token usage without requiring long intermediate generations. A key component is stability-based adaptive halting, which monitors anchor dynamics and terminates refinement once updates converge, enabling instance-wise allocation of latent computation under a shared maximum budget.

Empirically, AdaAnchor improves the efficiency--accuracy trade-off relative to fixed-step latent refinement while also substantially reducing output-token usage compared to token-level reasoning baselines. Under the same maximum latent budget, adaptive halting improves accuracy by up to $\sim$5\% over fixed-step refinement while reducing the average number of latent refinement iterations by $\sim$48--60\%. In addition, by emitting answer-only outputs and performing computation silently in latent space, AdaAnchor reduces generated tokens by $\sim$92--93\% compared to token-level reasoning baselines, illustrating how latent refinement can support iterative reasoning with fewer generated tokens and fewer refinement steps.

\bibliography{iclr2026_conference}
\bibliographystyle{iclr2026_conference}


\end{document}

%% file: math_commands.tex
\usepackage{amsmath,amsfonts,bm}

\def\eqref#1{equation~\ref{#1}}

\def\1{\bm{1}}

\DeclareMathAlphabet{\mathsfit}{\encodingdefault}{\sfdefault}{m}{sl}
\SetMathAlphabet{\mathsfit}{bold}{\encodingdefault}{\sfdefault}{bx}{n}

